\documentclass[conference,a4paper]{IEEEtran}
\IEEEoverridecommandlockouts
\usepackage[dvips]{graphicx}  

\usepackage{multirow}
\usepackage[left=0.71in,top=0.94in,right=0.71in,bottom=1.18in]{geometry}
\begin{document}
\title{\huge Field Geology with a Wearable Computer:\\ First Results of the Cyborg Astrobiologist System$^{*}$\footnoterule\thanks{$^{*}$This work is partially
supported by three Ramon y Cajal fellowships to P. McGuire, J. Orm\"o and E. D\'iaz Mart\'inez, and by a grant to J. Orm\"o (\#AYA2003-01203), all from the Spanish Ministry for Science and Technology . The equipment used in this work was purchased by grants to our Center for Astrobiology from its sponsoring research organizations, CSIC and INTA.}}

\author{\authorblockN{Patrick C. McGuire$^1$\footnoterule\thanks{$^1$Corresponding author, Email: mcguire@physik.uni-bielefeld.de}, Javier G\'omez Elvira,\\ Jos\'e Antonio Rodr\'iguez Manfredi, and Eduardo Sebasti\'an Mart\'inez}
\authorblockA{
{\it Robotics Laboratory, Centro de Astrobiolog\'ia (CAB)}\\
{\it Carretera de Torrej\'on a Ajalvir km 4.5}\\
{\it Torrej\'on de Ardoz, Madrid, Spain 28850}\\}
\and 
\authorblockN{Jens Orm\"o}
\authorblockA{
{\it Planetary Geology Laboratory}\\
{\it Centro de Astrobiolog\'ia (CAB)}\\
{\it Carretera de Torrej\'on a Ajalvir km 4.5}\\
{\it Torrej\'on de Ardoz, Madrid, Spain 28850}\\}
\and
\authorblockN{Enrique D\'iaz Mart\'inez$^2$\footnoterule\thanks{$^2$(formerly at) Planetary Geology Laboratory, CAB}}
\authorblockA{
{\it Direcci\'on de Geolog\'ia y Geof\'isica}\\
{\it Instituto Geol\'ogico y Minero de Espa\~na}\\
{\it Calera 1, Tres Cantos, Madrid, Spain 28760}\\}
\and
\authorblockN{Helge Ritter, Markus Oesker, Robert Haschke and J\"org Ontrup}
\authorblockA{
{\it Neuroinformatics Group, Computer Science Department}\\
{\it Technische Fakult\"at, University of Bielefeld}\\
{\it P.O.-Box 10 01 31; Bielefeld, Germany 33501}
}
}
\maketitle
\begin{abstract}
We present results from the first geological field tests of the `Cyborg Astrobiologist', which is a wearable computer and video camcorder system that we are using to test and train a computer-vision system towards having some of the autonomous decision-making capabilities of a field-geologist. The Cyborg Astrobiologist platform has thus far been used for testing and development of these algorithms and systems: robotic acquisition of quasi-mosaics of images, real-time image segmentation, and real-time determination of interesting points in the image mosaics.  The hardware and software systems function reliably, and the computer-vision algorithms are adequate for the first field tests. In addition to the proof-of-concept aspect of these field tests, the main result of these field tests is the enumeration of those issues that we can improve in the future, including: dealing with structural shadow and microtexture, and also, controlling the camera's zoom lens in an intelligent manner. Nonetheless, despite these and other technical inadequacies, this Cyborg Astrobiologist system, consisting of a camera-equipped wearable-computer and its computer-vision algorithms, has demonstrated its ability of finding genuinely interesting points in real-time in the geological scenery, and then gathering more information about these interest points in an automated manner.
  \end{abstract}

\section{Introduction}

Outside of the Mars robotics community, it is commonly presumed that
the robotic rovers on Mars are controlled in a time-delayed joystick manner, wherein commands are sent to the rovers several if not many times per day, as new information is acquired from the rovers' sensors.  However, inside the Mars robotics community, they have learned that such a brute force joystick-control process is rather cumbersome, and they have developed much more elegant methods for robotic control of the rovers on Mars, with highly significant degrees of robotic autonomy.

 Particularly, the Mars Exploration Rover (MER) team has demonstrated autonomy for the two robotic rovers Spirit \& Opportunity to the level that: practically all commands for a given Martian day (1 `sol' $=$ 24.6 hours) are delivered to each rover from Earth before the robot wakens from its power-conserving nighttime resting mode \cite{Crisp}\cite{Squyres}. Each rover then follows the commanded sequence of moves for the entire sol, moving to desired locations, articulating its arm with its sensors to desired points in the workspace of the robot, and acquiring data from the cameras and chemical sensors. From an outsider's point of view, these capabilities may not seem to be significantly autonomous, in that all the commands are being sent from Earth, and the MER rovers are merely executing those commands. But the following facts/feats deserve emphasis before judgement is made of the quality of the MER autonomy: this robot is on another planet with a complex surface to navigate and study; and all of the complex command sequence is sent to the robot the previous night for autonomous operation the next day. Sophisticated software and control systems are also part of the system, including the MER autonomous obstacle avoidance system and the MER visual odometry \& localization software. One should remember that there is a large team of human roboticists and geologists working here on the Earth in support of the MER missions, to determine science targets and robotic command sequences on a daily basis; currently, after the sun sets for an MER rover, the rover mission team can determine the science priorities and the command sequence for the next sol in about 4-5 hours.

 One future mission deserves special discussion for the technology developments described in this paper: the Mars Science Laboratory, planned for launch in 2009 (MSL'2009). A particular capability desired for this MSL'2009 mission will be to rapidly traverse to up to three geologically-different scientific points-of-interest within the landing ellipse.  These three geologically-different sites will be chosen from Earth by analysis of relevant satellite imagery. Possible desired maximal traversal rates could range from 300-2000 meters/sol in order to reach each of the three points-of-interest in the landing ellipse in minimum time.

Given these substantial expected traversal rates of the MSL'2009 rover, autonomous obstacle avoidance \cite{Goldberg} and autonomous visual odometry \& localization \cite{Olson} will be essential to achieve these rates, since otherwise, rover damage and slow science-target approach would be the results.  Given such autonomy in the rapid traverses, it behooves us to enable the autonomous rover with sufficient scientific responsibility. Otherwise, the robotic rover exploration system might drive right past an important scientific target-of-opportunity along the way to the human-chosen scientific point-of-interest. Crawford \& Tamppari \cite{Crawford} and their NASA/Ames team summarize possible `autonomous traverse science', in which every 20-30 meters during a 300 meter traverse (in their example), science pancam and Mini-TES (Thermal Emission Spectrometer) image mosaics are autonomously obtained. They state that ``there {\it may be} onboard analysis of the science data from the pancam and the mini-TES, which compares this data to predefined signatures of carbonates or other targets of interest. If detected, traverse may be halted and information relayed back to Earth.'' This onboard analysis of the science data is precisely the technology issue that we have been working towards solving. This paper is the first report to the general robotics community describing our progress towards giving a robotic astrobiologist some aspects of autonomous recognition of scientific targets-of-opportunity. 

 Before proceeding, we first note here two of the related efforts in the development of autonomous recognition of scientific targets-of-opportunity for astrobiological exploration: firstly, the work on developing a Nomad robot to search for meteorites in Antartica led by the Carnegie Mellon University Robotics Institute \cite{Apostolopoulos}\cite{Pederson}, and secondly, the work by a group at NASA/Ames on developing a Geological Field Assistant (GFA) \cite{Gulick1}\cite{Gulick2}\cite{Gulick3}.  

 \begin{figure}[h]
 \includegraphics[width=3in]{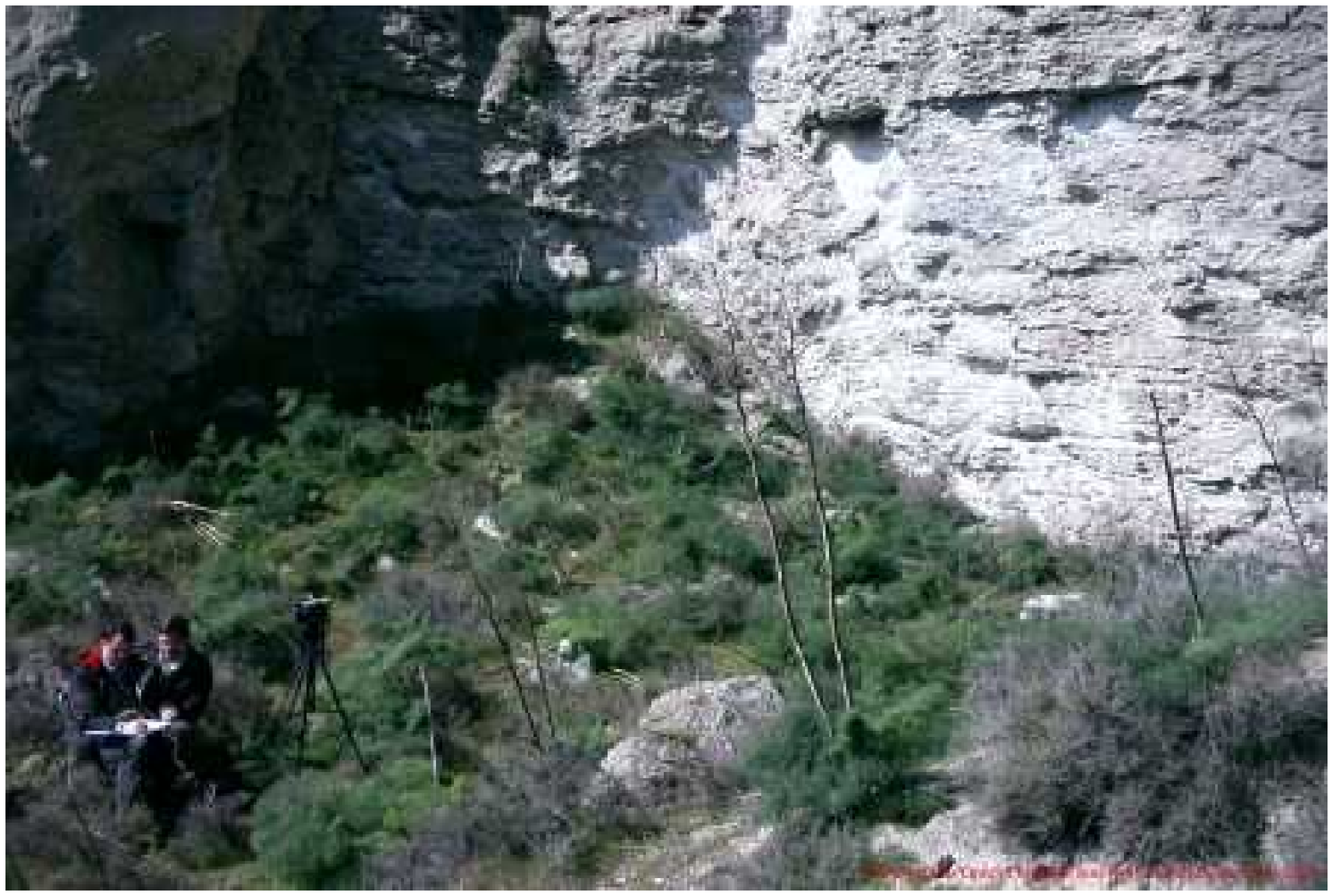}
  \caption{ D\'iaz Mart\'inez \& McGuire with the Cyborg Astrobiologist
System on 3 March 2004, 10 meters from the outcrop cliff
that is being studied during the first geological field mission near Rivas Vaciamadrid. We are taking notes
prior to acquiring one of our last-of-the-day mosaics and its set of interest-point chips.
Photo copyright: D\'iaz Mart\'inez, Orm\"o \& McGuire }
 \end{figure}

\section{The Cyborg Geologist \& Astrobiologist System}
 Our ongoing effort in this area of autonomous recognition of scientific targets-of-opportunity for field geology and field astrobiology is beginning to mature as well. To date, we have developed and field-tested a GFA-like ``Cyborg Astrobiologist'' system \cite{McGuire1}\cite{McGuire2} that now can:
\begin{trivlist}
\raggedright
\item $\bullet$ Use human mobility to maneuver to and within a geological site and to follow suggestions from the computer as to how to approach a geological outcrop;
\item $\bullet$ Use a portable robotic camera system to obtain a mosaic of color images;
\item $\bullet$ Use a `wearable' computer to search in real-time for the most uncommon regions of these mosaic images;
\item $\bullet$ Use the robotic camera system to re-point at several of the most uncommon areas of the mosaic images, in order to obtain much more detailed information about these `interesting' uncommon areas;
\item $\bullet$ Use human intelligence to choose between the wearable computer's different options for interesting areas in the panorama for closer approach; and
\item $\bullet$ Repeat the process as often as desired, sometimes retracing a step of geological approach. 
\end{trivlist}

   In the Mars Exploration Workshop in Madrid in November 2003, we demonstrated some of the early capabilities of our `Cyborg' Geologist/Astrobiologist System\cite{McGuire1}. We have been using this Cyborg system as a platform to develop computer-vision algorithms for recognizing interesting geological and astrobiological features, and for testing these algorithms in the field here on the Earth.

   The half-human/half-machine `Cyborg' approach (Fig.~1) uses human locomotion and human-geologist intuition/intelligence for taking the computer vision-algorithms to the field for teaching and testing, using a wearable computer. This is advantageous because we can therefore concentrate on developing the `scientific' aspects for autonomous discovery of features in computer imagery, as opposed to the more `engineering' aspects of using computer vision to guide the locomotion of a robot through treacherous terrain. This means the development of the scientific vision system for the robot is effectively decoupled from the development of the locomotion system for the robot.

   After the maturation of the computer-vision algorithms, we hope to
transplant these algorithms from the Cyborg computer to the on-board
computer of a semi-autonomous robot that will be bound for Mars or one of the interesting moons in our solar system.

 \begin{figure}[th]
 \center{\includegraphics[width=3.0in]{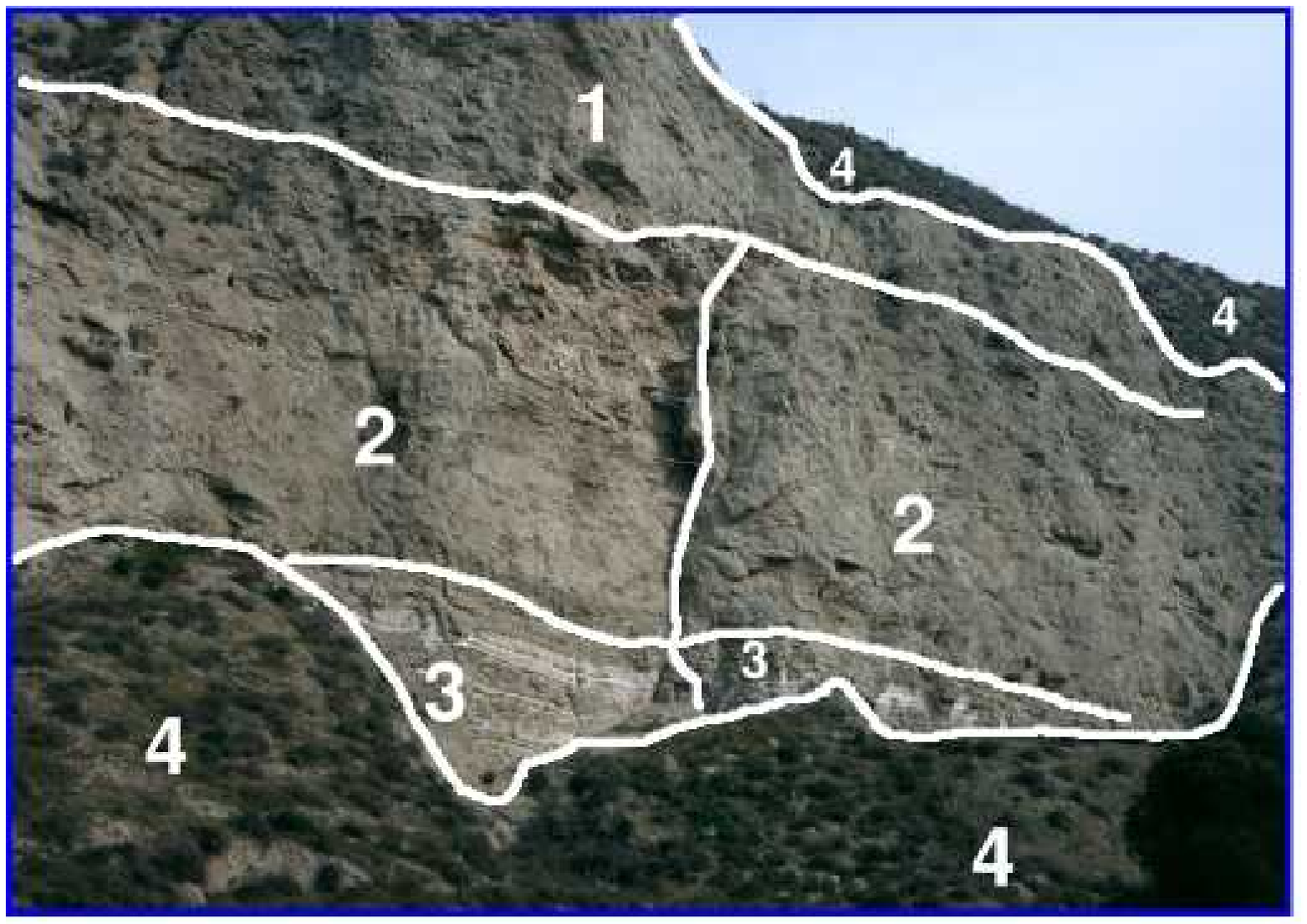}}
  \caption{An image segmentation made by human geologist D\'iaz Mart\'inez of the outcrop during the first mission to Rivas Vaciamadrid. Region 1 has a tan color and a blocky texture; Region 2 is subdivided by a vertical fault and has more red color and a more layered texture than Region 1; Region 3 is dominated by white and tan layering; and Region 4 is covered by vegetation. The dark \& wet spots in Region 3 were only observed during the second mission, 3 months later. The Cyborg Geologist/Astrobiologist made its own image segmentations for portions of the cliff face that included the area of the white layering at the bottom of the cliff (Fig.~5).  Photo copyright: D\'iaz Mart\'inez, Orm\"o \& McGuire}
 \end{figure}

Both of the field geologists on our team, D\'iaz Mart\'inez and Orm\"o, have independently stressed the importance to field geologists of geological `contacts' and the differences between the geological units that are separated by the geological contact. For this reason, in March 2003, we decided that the most important tool to develop for the beginning of our computer vision algorithm development was that of  `image segmentation'. Such image segmentation algorithms would allow the computer to break down a panoramic image into different regions (Fig.~2 for an example), based upon similarity, and to find the boundaries or contacts between the different regions in the image, based upon difference. Much of the remainder of this paper discusses the first geological field trials with the wearable computer of the segmentation algorithm we have developed over the last year.

\subsection{Image Segmentation, Uncommon Maps, Interest Maps, and Interest Points}

With human vision, a geologist:
\begin{trivlist}
\raggedright
\item$\bullet$ Firstly, tends to pay attention to those areas of a scene which are most unlike the other areas of the scene; and then,
\item$\bullet$ Secondly, attempts to find the relation between the different areas of the scene, in order to understand the geological history of the outcrop.
\end{trivlist}

The first step in this prototypical thought process of a geologist was our motivation for inventing the concept of uncommon maps. See Fig.~3 for a simple illustration of the concept of an uncommon map.
We have not yet attempted to solve the second step in this prototypical thought process of a geologist, but it is evident from the formulation of the second step, that
human geologists do not immediately ignore the common areas of the scene.  Instead, human geologists catalog the common areas and put them in the back of their minds for
``higher-level analysis of the scene'', or in other words, for determining explanations for the relations of the uncommon areas of the scene with the common areas of the scene.

 \begin{figure}[th]
 \center{\includegraphics[width=2in]{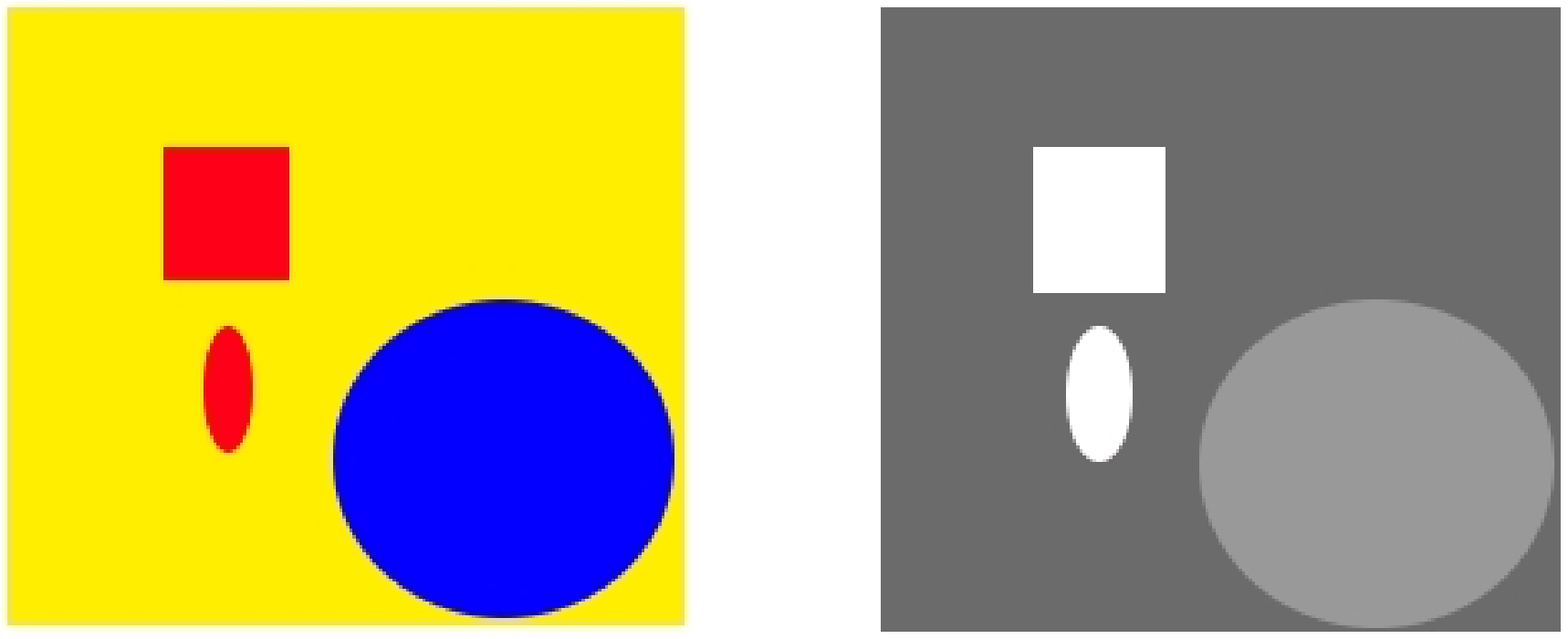}}
  \caption{For the simple, idealized image on the left, we show the corresponding uncommon map on the right. The whiter areas in the uncommon map are more uncommon than the darker areas in this map. }
 \end{figure}

 Prior to implementing the `uncommon map', the first step of the prototypical geologist's thought process, we needed a segmentation algorithm, in order to produce pixel-class maps to serve as input to the uncommon map algorithm.
We have implemented the classic co-occurrence histogram algorithm \cite{Haralick}\cite{Haddon}. For this work, we have not included texture information in the segmentation algorithm and the uncommon map algorithm. Currently, each of the three bands of $HSI$ color information is segmented separately, and later merged in the interest map by summing three independent uncommon maps. Maybe in future work, image segmentation simultaneously using color \& texture could be developed for and tested on the Cyborg Astrobiologist System (e.g., \cite{Freixenet}).

The concept of an `uncommon map' is our invention, though it probably has been independently invented by other authors, since it is somewhat useful. In our implementation, the uncommon map algorithm takes the top 8 pixel classes determined by the image segmentation algorithm, and ranks each pixel class according to how many pixels there are in each class.  The pixels in the pixel class with the greatest number of pixel members are numerically labelled as `common', and the pixels in the pixel class with the least number of pixel members are numerically labelled as 'uncommon'.  The `uncommonness' hence ranges from 1 for a common pixel to 8 for an uncommon pixel, and we can therefore construct an uncommon map given any image segmentation map.  In our work, we construct several uncommon maps from the color image mosaic, and then we sum these uncommon maps together, in order to arrive at a final interest map. 

In this paper, we develop and test a simple, high-level concept of interest points of an image, which is based upon finding the centroids of the smallest (most uncommon) regions of the image. Such a `global' high-level concept of interest points differs from the lower-level `local' concept of F\"orstner interest points based upon corners and centers of circular features. However, this latter technique with local interest points is used by the MER team for their stereo-vision image matching and for their visual-odometry and visual-localization image matching \cite{Goldberg}\cite{Olson}\cite{Nesnas}. Our interest point method bears somewhat more relation to the higher-level wavelet-based salient points technique \cite{Sebe}, in that they search first at coarse resolution for the image regions with the largest gradient, and then they use wavelets in order to zoom in towards the salient point within that region that has the highest gradient. Their salient point technique is edge-based, whereas our interest point is currently region-based. Since in the long-term, we have an interest in geological contacts, this edge-based \& wavelet-based salient point technique could be a reasonable future interest-point algorithm to incorporate into our Cyborg Astrobiologist system for testing.

\subsection{Hardware \& Software  for the Cyborg Astrobiologist}

The non-human hardware of the Cyborg Astrobiologist system consists of:
\begin{trivlist}
\raggedright
\item $\bullet$ a 667 MHz wearable computer (from ViA Computer Systems) with a `power-saving' Transmeta `Crusoe' CPU and 112 MB of physical memory, 
\item $\bullet$ an {\it SV-6} Head Mounted VGA Display (from Tekgear, via the Spanish supplier Decom) that works well in bright sunlight,
\item $\bullet$ a SONY `Handycam' color video camera (model {\it DCR-TRV620E-PAL}), with a Firewire/IEEE1394 cable to the computer,
\item $\bullet$ a thumb-operated USB finger trackball from 3G Green Green Globe Co., resupplied by ViA Computer Systems and by Decom,
\item $\bullet$ a small keyboard attached to the human's arm,
\item $\bullet$ a tripod for the camera, and
\item $\bullet$ a Pan-Tilt Unit (model {\it PTU-46-70W}) from Directed Perception with a bag of associated power and signal converters.
\end{trivlist}

 The wearable computer processes the images acquired by the color digital video camera, to compute a map of interesting areas. The computations include: simple mosaicking by image-butting, as well as two-dimensional histogramming for image segmentation \cite{Haralick}\cite{Haddon}.  This image segmentation is independently computed for each of the Hue, Saturation, and Intensity (H,S,I) image planes, resulting in three different image-segmentation maps. These image-segmentation maps were used to compute `uncommon' maps (one for each of the three (H,S,I) image-segmentation maps): 
each of the three resulting uncommon maps gives highest weight to those regions of smallest area for the respective (H,S,I) image planes. Finally, the
three (H,S,I) uncommon maps are added together into an interest map, which is used by the Cyborg system for subsequent interest-guided pointing of the camera.

  After segmenting the mosaic image (Fig.~5), it becomes obvious that a very simple method to find interesting regions in an image is to look for those regions in the image that have a significant number of uncommon pixels. We accomplish this by (Fig.~6): first, creating an uncommon map based upon a linear reversal of the segment area ranking; second, adding the 3 uncommon maps (for $H$, $S$, \& $I$) together to form an interest map; and third, blurring this interest map\footnote{with a gaussian smoothing kernel of width $B=10$ pixels.}.

  Based upon the three largest peaks in the blurred/smoothed interest map, the Cyborg system then guides the Pan-Tilt Unit to point the camera at each of these three positions to acquire high-resolution color images of the three interest points (Fig.~4). By extending a simple image-acquisition and image-processing system to include robotic and mosaicking elements, we were able to conclusively demonstrate that the system can make reasonable decisions by itself in the field for robotic pointing of the camera.

\section{Descriptive Summaries of the Field Site and of the Expeditions}

   On the March 3rd and June 11th, 2004, three of the authors (McGuire, D\'iaz Mart\'inez \& Orm\"o) tested the ``Cyborg Astrobiologist" system for the first time at a geological site, the gypsum-bearing southward-facing stratified cliffs near the ``El Campillo" lake of Madrid's Southeast Regional Park, outside the suburb of Rivas Vaciamadrid. Due to the significant storms in the 3 months between the two missions, there were 2 dark \& wet areas in the gypsum cliffs that were visible only during the second mission.  In Fig.~2, we show the segmentation of the outcrop (during the first mission), according to the human Geologist D\'iaz Martinez, for reference.

 The computer was worn on McGuire's belt, and typically took 3-5 minutes to acquire and compose a mosaic image composed of $M \times N$ subimages. Typical values of $M \times N$ used were $3 \times 9$ and $11 \times 4$. The sub-images were downsampled in both directions by a factor of 4-8 during these tests; the original sub-image dimensions were $360 \times 288$.

   Several mosaics were acquired of the cliff face from a distance of about 300 meters, and the computer automatically determined the three most interesting points in each mosaic. Then, the wearable computer automatically repointed the camera towards each of the three interest points, in order to acquire {\it non-downsampled} color images of the region around each interest point in the image. All the original mosaics, all the derived mosaics and all the interest-point subimages were then saved to hard disk for post-mission study.

 Two other tripod positions were chosen for acquiring mosaics and interest-point image-chip sets. At each of the three tripod positions, 2-3 mosaic images and interest-point image-chip sets were acquired. One of the chosen tripod locations was about 60 meters from the cliff's face; the other was about 10 meters (Fig.~1) from the cliff face.

During the 2nd mission at distances of 300 meters and 60 meters, the system most often determined the wet spots (Fig.~4) to be the most interesting regions on the cliff face. This was encouraging to us, because we also found these wet spots to be the most interesting regions\footnote{These dark \& wet regions were interesting to us partly because they give information about the development of the outcrop. Even if the relatively small spots were only dark, and not wet (i.e., dark dolerite blocks, or a brecciated basalt), their uniqueness in the otherwise white \& tan outcrop would have drawn our immediate attention. Additionally, even if this had been our first trip to the site, and if the dark spots had been present during this first trip, these dark regions would have captured our attention for the same reasons. The fact that these dark spots had appeared after our first trip and before the second trip was not of paramount importance to grab our interest (but the `sudden' appearance of the dark spots between the two missions did arouse our higher-order curiosity).}.

After the tripod position at 60 meters distance, we chose the next tripod position to be about 10 meters from the cliff face (Fig.~1).  During this `close-up' study of the cliff face, we intended to focus the Cyborg Astrobiologist exploration system upon the two points that it found most interesting when it was in the more distant tree grove, namely the two wet and dark regions of the lower part of the cliff face.  By moving from 60 meters distance to 10 meters distance and by focusing at the closer distance on the interest points determined at the larger distance, we wished to simulate how a truly autonomous robotic system would approach the cliff face.  Unfortunately, due to a combination of a lack of human foresight in the choice of tripod position and a lack of more advanced software algorithms to mask out the surrounding \& less interesting region (see discussion in Section 4), for one of the two dark spots, the Cyborg system only found interesting points on the undarkened periphery of the dark \& wet stains. Furthermore, for the other dark spot, the dark spot was spatially complex, being subdivided into several regions, with some green and brown foliage covering part of the mosaic. Therefore, in both close-up cases the value of the interest mapping is debatable. This interest mapping could be improved in the future, as we discuss in Section 4.2.

\section{Results}  
\subsection{Results from the First Geological Field Test}

As first observed during the first mission to Rivas on March 3rd, the characteristics of the southward-facing cliffs at Rivas Vaciamadrid consist of mostly tan-colored surfaces, with some white veins or layers, and with significant shadow-causing three-dimensional structure. The computer vision algorithms performed adequately for a first visit to a geological site, but they need to be improved in the future. As decided at the end of the first mission by the mission team, the improvements include: shadow-detection and shadow-interpretation algorithms,
and segmentation of the images based upon microtexture.

 In the last case, we decided that due to the very monochromatic \& slightly-shadowy nature of the imagery, the Cortical Interest Map algorithm non-intuitively decided to concentrate its interest on differences in intensity, and it tended to ignore hue and saturation. 

After the  first geological field test, we spent several months studying the imagery obtained during this mission, and fixing various further problems that were only discovered after the first mission. Though we had hoped that the first mission to Rivas would have been more like a science mission, in reality it was more of an engineering mission.

\subsection{Results from the Second Geological Field Test}

In Fig.~4, from the tree grove at a distance of 60 meters, the Cyborg Astrobiologist system found the dark \& wet spot on the right side to be the most interesting, the dark \& wet spot on the left side to be the second most interesting, and the small dark shadow in the upper left hand corner to be the 3rd most interesting. For the first two  interest points (the dark \& wet spots), it is apparent from the uncommon map for intensity pixels in Fig.~6 that these points are interesting due to their relatively remarkable intensity values. By inspection of Fig.~5, we see that these pixels which reside in the white segment of the intensity segmentation mosaic are unusual because they are a cluster of very dim pixels (relative to the brighter red, blue and green segments). Within the dark wet spots,  we observe that these particular points in the white segment of the intensity segmentation in Fig.~5 are interesting because they reside in the {\it shadowy} areas of the dark \& wet spots. We interpret the interest in the 3rd interest point to be due to the juxtaposition of the small green plant with the shadowing in this region; the interest in this point is significantly smaller than for the 2 other interest points.

More advanced software could be developed to handle better the close-up real-time interest-map analysis of the imagery acquired at the close-up tripod position (10 meter distance from the cliff; not shown here). Here are some options to be included in such software development:
\begin{trivlist}
\raggedright
\item $\bullet$ Add hardware \& software to the Cyborg Astrobiologist so that it can make intelligent use of its zoom lens. We would need to study \& develop the camera's LANC communication interface, or possibly determine if the zoom lens could be controlled over its Firewire/IEEE1394 communication cable. With such software for intelligent zooming, the system could have corrected the human's mistake in tripod placement and decided to zoom further in, to focus only on the shadowy part of the dark \& wet spot (which was determined to be the most interesting point at a distance of 60 meters), rather than the periphery of the entire dark \& wet spot. 
\item $\bullet$ Enhance the Cyborg Astrobiologist system so that it has a memory of the image segmentations performed at a greater distance or at a lower magnification of the zoom lens. Then, when moving to a closer tripod position or a higher level of zoom-magnification, register the new imagery or the new segmentation maps with the coarser resolution imagery and segmentation maps. Finally, tell the system to mask out or ignore or deemphasize those parts of the higher resolution imagery which were part of the low-interest segments of the coarser, more distant segmentation maps, so that it concentrates on those features that it determined to be interesting at coarse resolution and higher distance. 
\end{trivlist}

\begin{figure}[h]

\center{\includegraphics[width=3.2in]{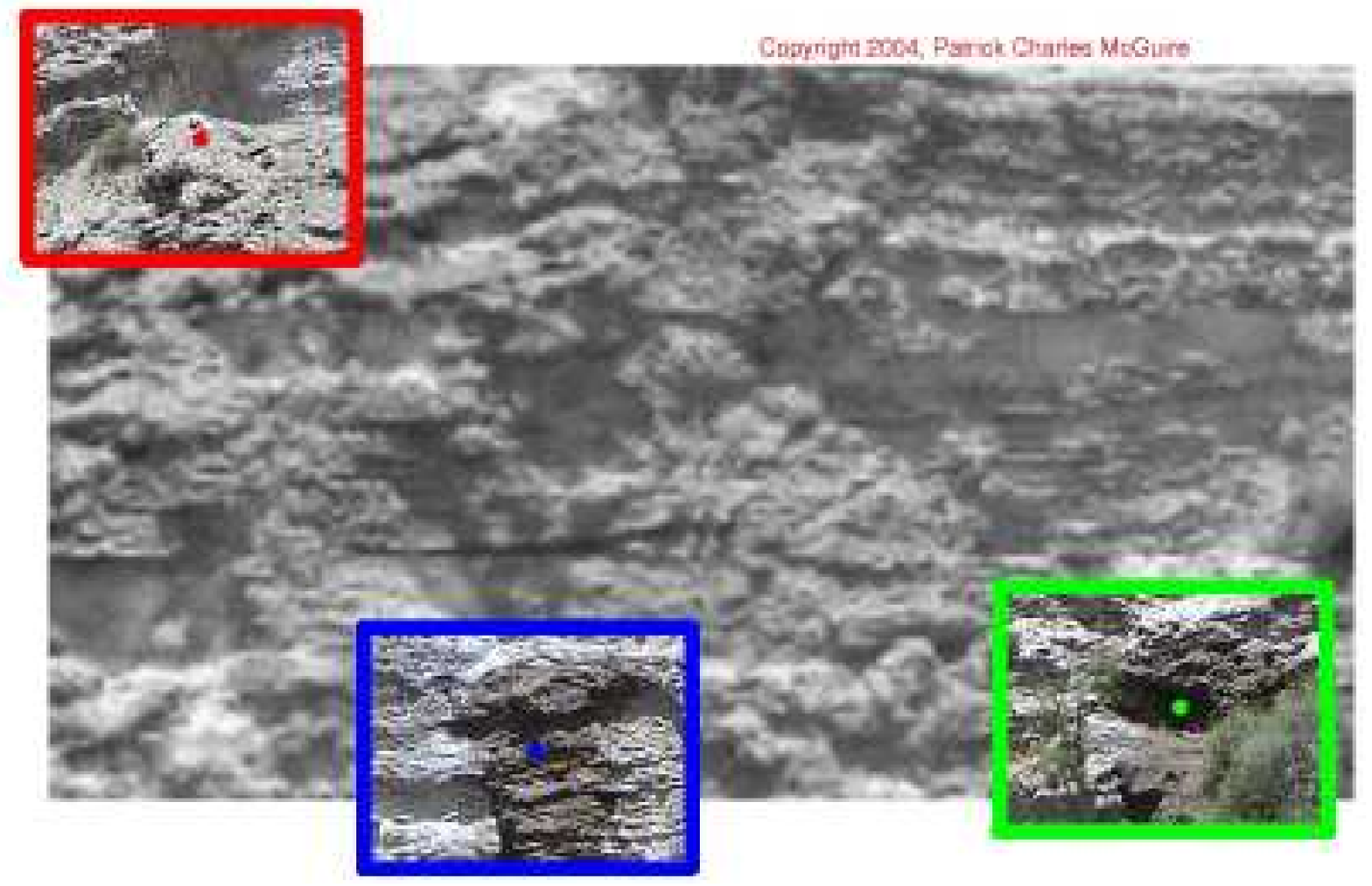}}

\caption{Mosaic image of a three-by-four set of grayscale sub-images acquired by the 
Cyborg Astrobiologist at the beginning of its second expedition.
The three most interesting points were subsequently revisited by the
camera in order to acquire full-color higher-resolution images of
these points-of-interest.  
 The colored points and rectangles represent the points that the Cyborg
 Astrobiologist determined (on location) to be most interesting;
{\it green} is most interesting, {\it blue} is second most interesting, and {\it red} is third most interesting.
    The images were taken and processed in real-time between 1:25PM and 1:35PM local time
on 11 June 2004 about 60 meters from some gypsum-bearing southward-facing
cliffs near the ``El Campillo" lake of the Madrid southeast regional park
outside of Rivas Vaciamadrid. See Figs.~5~\&~6 for some details about the real-time image processing that was done in order to determine the location of the interest points in this figure.
 }

 \end{figure}

\section{Discussion \& Conclusions}

   Both the human geologists on our team concur with the judgement of the Cyborg Astrobiologist software system, that the two dark \& wet spots on the cliff wall were the most interesting spots during the second mission. However, the two geologists also state that this largely depends on the aims of study for the geological field trip; if the aim of the study is to search for hydrological features, then these two dark \& wet spots are certainly interesting. One question which we have thus far left unstudied is ``What would the Cyborg Astrobiologist system have found interesting during the second mission if the two dark \& wet spots had not been present during the second mission?'' It is possible that it would again have found some dark shadow particularly interesting, but with the improvements made to the system between the first and second mission, it is also possible that it could have found a different feature of the cliff wall more interesting.

\subsection{Outlook}

   The NEO programming for this Cyborg Geologist project was initiated with the SONY Handycam in April 2002. The wearable computer arrived in June 2003, and the head mount\-ed display arrived in November 2003. We now have a reliably functioning human and hardware and software Cyborg Geologist system, which is partly robotic with its Pan Tilt camera mount. This robotic extension allows the camera to be pointed repeatedly, precisely \& automatically in different directions.

      Based upon the significantly-improved performance of the Cyborg Astrobiologist system during the 2nd mission to Rivas in June 2004, we conclude that the system now is debugged sufficiently so as to be able to produce studies of the utility of particular computer vision algorithms for geological deployment in the field. We have outlined some possibilities for improvement of the system based upon the second field trip, particularly in the improvement in the systems-level algorithms needed in order to more intelligently drive the approach of the Cyborg or robotic system towards a complex geological outcrop. These possible systems-level improvements include: hardware \& software for intelligent use of the camera's zoom lens and a memory of the image segmentation performed at greater distance or lower magnification of the zoom lens.

\section{Acknowledgements}

We thank certain individuals for assistance or conversations: 
Virginia Souza-Egipsy, Mar\'ia Paz Zorzano Mier, Carmen C\'ordoba Jabonero, Kai Neuffer, Antonino Giaquinta, Fernando Camps Mart\'inez, Alain Lepinette Malvitte, Josefina Torres Redondo, V\'ictor R. Ruiz, Julio Jos\'e Romeral Planell\'o, Gemma Delicado, Jes\'us Mart\'inez Fr\'ias, Irene Schneider, Gloria Gallego, Carmen Gonz\'alez, Ramon Fern\'andez, Coronel Angel Santamaria, Carol Stoker, Paula Grunthaner, Maxwell D. Walter, Fernando Ayll\'on Quevedo, Javier Mart\'in Soler, Juan P\'erez Mercader, J\"org Walter, Claudia Noelker, Gunther Heidemann, Robert Rae, and Jonathan Lunine. 


 \begin{figure}[h]

\center{\includegraphics[height=2.8in]{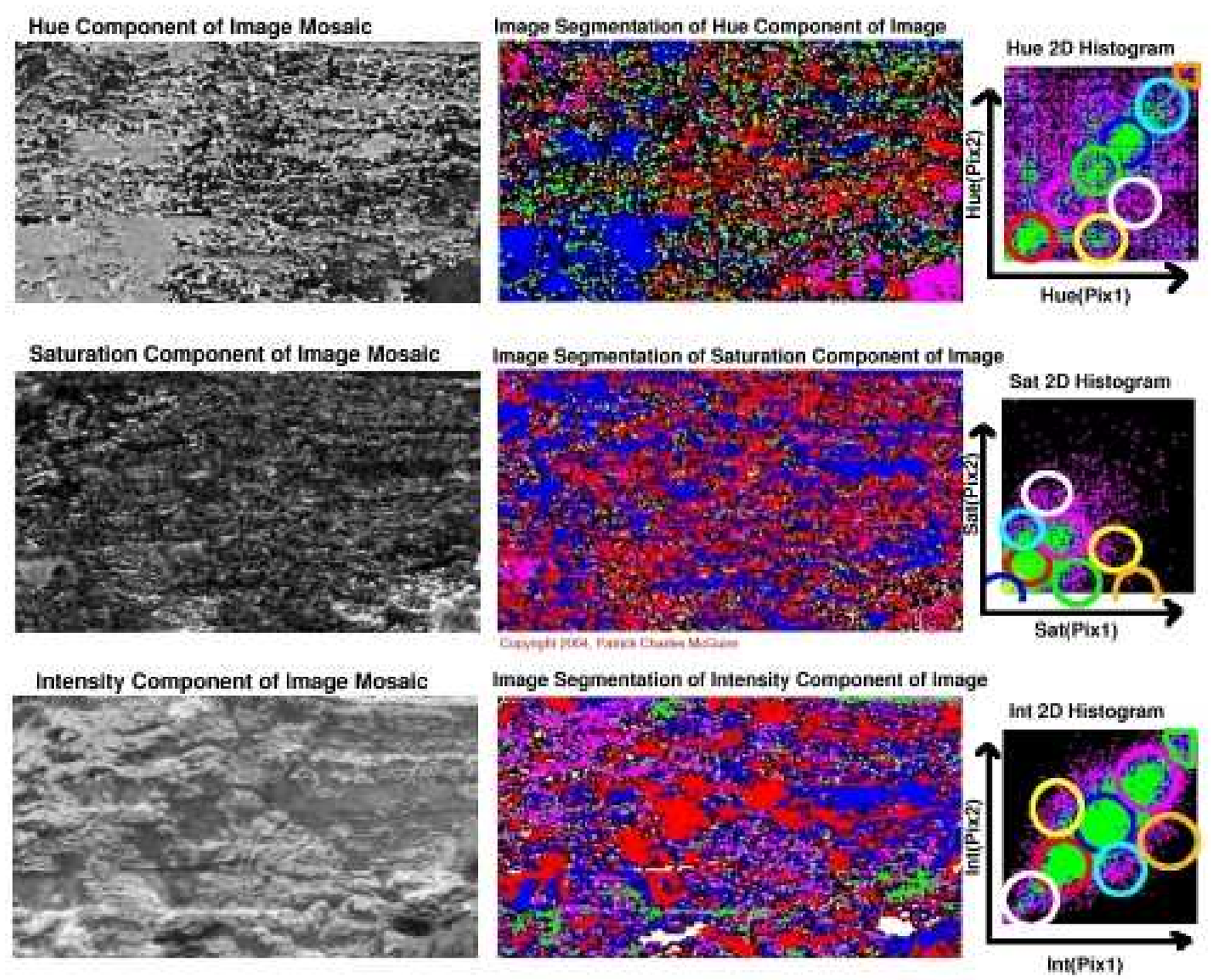}}

\caption{
   In the middle column, we show the three image-segmentation maps computed in real-time by the Cyborg Astrobiologist system, based upon the original Hue, Saturation \& Intensity ($H$, $S$ \& $I$) mosaics in the left column and the derived 2D co-occurrence histograms shown in the right column. The wearable computer made this and all other computations for the original $3\times4$ mosaic ($108\times192$ pixels, shown in Fig.~4) in about 2 minutes after the initial acquisition of the mosaic sub-images was completed.
   The colored regions in each of the three image-segmentation maps
correspond to pixels \& their neighbors in that map that have similar
statistical properties in their two-point correlation values, as shown by the circles of corresponding colors in the 2D histograms in the column on the right. 
         The RED-colored regions in the segmentation maps correspond to the mono-statistical
              regions with the largest area in this mosaic image;
              the RED regions are the least "uncommon" pixels in the
              mosaic.
         The BLUE-colored regions correspond to the mono-statistical
              regions with the 2nd largest area in this mosaic image;
              the BLUE regions are the 2nd least "uncommon" pixels in the
              mosaic.
         And similarily for the PURPLE, GREEN, CYAN, YELLOW, WHITE, and ORANGE.
         The pixels in the BLACK regions have failed to be segmented by the segmentation algorithm.
}
\end{figure}

\begin{figure}[b]

\center{\includegraphics[height=3.2in]{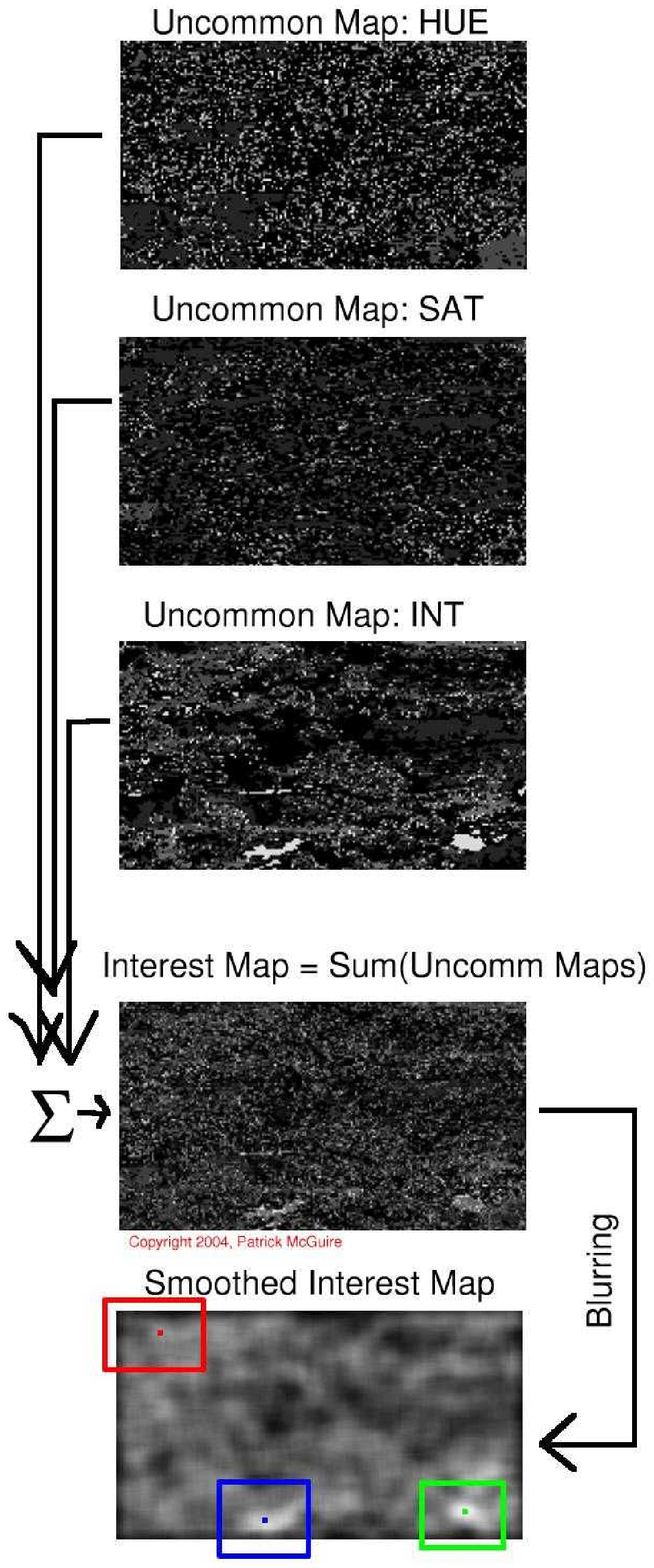}}

\caption{
  These are the uncommon maps for the mosaic shown in Fig.~4,  based on the region sizes determined by the image-segmentation algorithm shown in  Fig.~5.
  Also shown is the interest map, i.e., the unweighted sum of the three uncommon maps. We blur the original interest map before
determining the ``most interesting" points. These ``most interesting" points
are then sent to the camera's Pan/Tilt motor in order to acquire and
save-to-disk 3 higher-resolution RGB color images of the small areas in
the image around the interest points (Fig.~4).
   Green is the most interesting point. Blue is 2nd most interesting.
And Red is 3rd most interesting.
}

 \end{figure}


\begin{thebibliography}{99}
 

\bibitem{Crisp} 
J.~A. Crisp, M. Adler, J.~R. Matijevic, S.~W. Squyres, R.~E. Arvidson, 
\& D.~M. Kass, ``Mars Exploration Rover mission,''  {\it Journal of 
Geophysical Research (Planets)}, vol. 108, no. 2, p. 1 (2003). 

\bibitem{Squyres} 
S.~W. Squyres, R.~E. Arvidson, J.~F. Bell, {\it et al.}, ``The Spirit rover's Athena science investigation at Gusev Crater, Mars,''  {\it Science}, vol. 305, pp. 794-800 (2004). 

\bibitem{Goldberg} 
S.B. Goldberg, M.W. Maimone, \& L. Matthies, ``Stereo vision and rover navigation software for planetary exploration,'' {\it 2002 IEEE Aerospace Conference Proceedings}, Big Sky, Montana, vol. 5, pp. 2025-2036 (2002).

\bibitem{Olson}
C.F. Olson, L.H. Matthies, M. Schoppers, \& M.W. Maimone, ``Rover navigation using stereo ego-motion,''  {\it Robotics and Autonomous Systems}, vol. 43, no. 4, pp. 215-229 (2003). 

\bibitem{Crawford}
J. Crawford \& L.K. Tamppari, ``Mars Science Laboratory -- autonomy requirements analysis,'' unpublished (2002).

\bibitem{Apostolopoulos}
D. Apostolopoulos, M.D. Wagner, B. Shamah, L. Pedersen, K. Shillcutt, \& W.L. Whittaker, 
``Technology and field demonstration of robotic search for Antarctic meteorites,'' {\it International Journal of Robotics Research}, vol. 19, no. 11, pp. 1015-1032 (2000). 

\bibitem{Pederson}
L. Pedersen, ``Autonomous characterization of unknown environments,'' {\it 2001 IEEE International Conference on Robotics and Automation}, vol. 1, pp. 277-284 (2001). 

\bibitem{Gulick1} 
V.~C. Gulick, R.~L. Morris, M.~A. Ruzon, \& T.~L. Roush, 
``Autonomous image analyses during the 1999 Marsokhod rover field test,''
{\it Journal of Geophysical Research}, vol. 106, pp. 7745-7764 (2001). 

\bibitem{Gulick2} 
V.~C. Gulick, R.~L. Morris, J. Bishop, P. Gazis, R. Alena, \& 
M. Sierhuis, ``Geologist's Field Assistant: developing image and 
spectral analyses algorithms for remote science exploration,''  {\it Lunar 
and Planetary Institute Conference Abstracts}, vol. 33, p. 1961 (2002). 

\bibitem{Gulick3} 
V.~C. Gulick, S.~D. Hart, X. Shi, \& V.~L. Siegel, ``Developing 
an automated science analysis system for Mars surface exploration for MSL 
and beyond,''  {\it Lunar and Planetary Institute Conference Abstracts}, vol. 
35, p. 2121 (2004).  

\bibitem{McGuire1}
P.~C. McGuire, J.~A. Rodriguez Manfredi, E. Sebastian Martinez, J. Gomez Elvira, E. Diaz Martinez, J. Orm\"o, K. Neuffer, A. Giaquinta, F. Camps Martinez, A. Lepinette Malvitte, J. Perez Mercader, H. Ritter, M. Oesker, J. Ontrup, \& J. Walter, ``Cyborg systems as platforms for computer-vision algorithm-development for astrobiology,'' {\it Proceedings of the III European Workshop on Exo/Astrobiology}, held at the Centro de Astrobiologia, Madrid, http://arxiv.org/abs/cs.CV/0401004, {\it ESA SP}, vol. 545,  pp. 141-144 (2004). 

\bibitem{McGuire2}
P.~C. McGuire, J.~O. Orm\"o, E. Diaz Martinez, J.A. Rodriguez Manfredi, J. Gomez Elvira, H. Ritter, M. Oesker, \& J. Ontrup, ``The Cyborg Astrobiologist: first field experience,'' (submitted to {\it International Journal of Astrobiology}, August 2004, unpublished).

\bibitem{Haralick}
R.M. Haralick, K. Shanmugan, \& I. Dinstein, ``Texture features for image classification,'' {\it IEEE SMC-3} (6), pp. 610-621 (1973).

\bibitem{Haddon}
Haddon, J.F. \& J.F. Boyce,
``Image segmentation by unifying region and boundary information,''
{\it IEEE Trans. Pattern Anal.\ Mach.\ Intell.}, vol. 12, no. 10, pp. 929-948 (1990).

\bibitem{Freixenet}
J. Freixenet, X. Mu\~noz, J. Mart\'i, \& X. Llad\'o, ``Color texture segmentation by region-boundary cooperation,'' {\it Computer Vision -- ECCV 2004, Eighth European Conference on Computer Vision, Proceedings, Part II, Lecture Notes in Computer Science}, Springer, Prague, Czech Republic, Eds. Tom\'as Pajdla, Jir\'i Matas, vol. 3022, pp. 250-261 (2004). 
(Also available in the {\it CVonline} archive).

\bibitem{Nesnas}
I. Nesnas, M. Maimone, H. Das, ``Autonomous vision-based manipulation from a rover platform,'' {\it Proceedings of the CIRA Conference}, Monterey, California (1999).

\bibitem{Sebe}
N. Sebe, Q. Tian, E. Loupias, M. Lew, \& T.S. Huang, ``Evaluation of salient points techniques,'' {\it Image and Vision Computing, Special Issue on Machine Vision}, vol. 21, pp. 1087-1095 (2003). 

\end{thebibliography}
\end{document}